\documentclass[sigconf]{acmart}
\AtBeginDocument{%
  }

\copyrightyear{2026}
\acmYear{2026}
\setcopyright{cc}
\setcctype{by-nc-nd}
\acmConference[WWW Companion '26]{Companion Proceedings of the ACM Web Conference 2026}{April 13--17, 2026}{Dubai, United Arab Emirates}
\acmBooktitle{Companion Proceedings of the ACM Web Conference 2026 (WWW Companion '26), April 13--17, 2026, Dubai, United Arab Emirates}
\acmPrice{}
\acmDOI{10.1145/3774905.3793128}
\acmISBN{979-8-4007-2308-7/2026/04}

\usepackage{enumitem} % Make sure this is in your preamble
\usepackage{float}
\usepackage{stfloats}
\usepackage{latexsym}
\usepackage{listings}
\usepackage{longtable}
\usepackage{multicol}
\usepackage{url,tcolorbox}
\usepackage{graphicx}
\usepackage{wrapfig}
\usepackage{natbib}
\usepackage{booktabs,fvextra}
\usepackage{comment}
\usepackage{subcaption}
\usepackage{verbatim}
\usepackage{longtable}
\usepackage{tikz}
\usepackage{array}
\usepackage{enumitem,verbatimbox}
\usepackage{xspace}
\usepackage{algorithm} 
\usepackage{algpseudocode}
\usepackage{xcolor}
\usepackage{hyperref}

\hypersetup{
  colorlinks=true,
  urlcolor=blue,   % standard hyperlink blue
}

\begin{document}

\title{\textsc{PeeriScope}: A Multi-Faceted Framework for Evaluating Peer Review Quality}


\author{Sajad Ebrahimi}
\authornote{Corresponding author. Email: \texttt{s.ebrahimi@utoronto.ca}}
\orcid{0009-0003-1630-3938}
\affiliation{%
  \institution{Reviewerly}
 \city{Toronto}
 \state{Ontario}
 \country{Canada}
  }
% \email{sajad@reviewer.ly}

\author{Soroush Sadeghian}
\orcid{0009-0005-2172-7617}
\affiliation{%
  \institution{Reviewerly}
 \city{Toronto}
 \state{Ontario}
 \country{Canada}
  }
% \email{soroushsa@reviewer.ly}

\author{Ali Ghorbanpour}
\orcid{0009-0006-7383-8105}
\affiliation{%
  \institution{Reviewerly}
 \city{Toronto}
 \state{Ontario}
 \country{Canada}
  }
% \email{aligh@reviewer.ly}

\author{Negar Arabzadeh}
\orcid{0000-0002-4411-7089}
\affiliation{%
  \institution{Reviewerly, UC Berkeley}
 \city{Berkeley}
 \state{California}
 \country{United States}
  }
% \email{negara@reviewer.ly}

\author{Sara Salamat}
\orcid{0009-0007-3676-6023}
\affiliation{%
  \institution{Reviewerly}
 \city{Toronto}
 \state{Ontario}
 \country{Canada}
  }
% \email{sara@reviewer.ly}

\author{Seyed Mohammad Hosseini}
\orcid{0009-0000-8197-6109}
\affiliation{%
  \institution{Reviewerly}
 \city{Toronto}
 \state{Ontario}
 \country{Canada}
  }
% \email{muhan@reviewer.ly}

\author{Hai Son Le}
\orcid{0009-0003-2240-0451}
\affiliation{%
  \institution{Reviewerly}
 \city{Toronto}
 \state{Ontario}
 \country{Canada}
  }
% \email{sonny@reviewer.ly}

\author{Mahdi Bashari}
\orcid{0000-0002-9211-5475}
\affiliation{%
  \institution{Reviewerly}
 \city{Toronto}
 \state{Ontario}
 \country{Canada}
  }
% \email{bashari@reviewer.ly}

\author{Ebrahim Bagheri}
\orcid{0000-0002-5148-6237}
\affiliation{%
  \institution{Reviewerly, University of Toronto}
 \city{Toronto}
 \state{Ontario}
 \country{Canada}
}
% \email{ebrahim.bagheri@utoronto.ca}

\renewcommand{\shortauthors}{Sajad Ebrahimi et al.}

\begin{abstract}
    The increasing scale and variability of peer review in scholarly venues has created an urgent need for systematic, interpretable, and extensible tools to assess review quality. We present \textsc{PeeriScope}, a modular platform that integrates structured features, rubric-guided large language model assessments, and supervised prediction to evaluate peer review quality along multiple dimensions. 
    Designed for openness and integration, \textsc{PeeriScope} provides both a public interface and a documented API, supporting practical deployment and research extensibility. The demonstration illustrates its use for reviewer self-assessment, editorial triage, and large-scale auditing, and it enables the continued development of quality evaluation methods within scientific peer review. \textsc{PeeriScope} is available both as a live demo\footnote{\url{https://app.reviewer.ly/app/peeriscope}} and via API services\footnote{ \url{https://github.com/Reviewerly-Inc/Peeriscope}}
\end{abstract}


\begin{CCSXML}
<ccs2012>
   <concept>
       <concept_id>10010147.10010178.10010179.10003352</concept_id>
       <concept_desc>Computing methodologies~Information extraction</concept_desc>
       <concept_significance>500</concept_significance>
       </concept>
 </ccs2012>
\end{CCSXML}

\ccsdesc[500]{Computing methodologies~Information extraction}


% %%
% %% Keywords. The author(s) should pick words that accurately describe
% %% the work being presented. Separate the keywords with commas.
\keywords{Peer review quality assessment, LLM-based evaluation, Scholarly peer-review analytics, Automated review evaluation}

\maketitle
\section{Introduction}

Peer review is a cornerstone of scholarly publishing, ensuring the quality and credibility of scientific communication.
However, the quality of reviews varies widely, and most venues lack standardized or scalable mechanisms to assess them.
As conferences and journals continue to grow, this inconsistency raises concerns about fairness, transparency, and trust in the evaluation process.
Recent advances in large language models (LLMs) have further transformed the peer-review landscape~\cite{taechoyotin2025remorautomatedpeerreview}. Given their ability to produce fluent and well-structured prose, LLMs are increasingly used to draft review reports (as witnessed in the recent ICLR 2026 drama) \cite{thakkar2025llmfeedbackenhancereview}. Recent studies show, however, that although LLMs can mimic the surface form of expert feedback, their critiques often lack  depth, domain-specific reasoning, and reliable factual grounding; they also struggle with providing actionable recommendations tailored to a paper's actual weaknesses~\cite{zhou-etal-2024-llm,lin2025breaking}. Importantly, evaluating these emerging LLM-assisted practices remains difficult because peer-review datasets are inherently sparse, fragmented across venues, and largely inaccessible due to confidentiality constraints. This lack of annotated, high-quality review data makes it challenging to benchmark LLM-generated reviews or to compare them meaningfully with human judgments. As such, these challenges underscore the need for scalable and interpretable frameworks capable of evaluating (human- and LLM-generated) peer-reviews.

A growing body of research has started exploring automated approaches to review analysis and generation. Recent studies have examined the politeness and engagement of reviews~\cite{bharti2024politepeer}, the prevalence of superficial or ``lazy'' reviewing~\cite{purkayastha2025lazyreview}, and biases such as institutional or gender disparities~\cite{tomkins2017reviewer, helmer2017gender}. Other work has investigated the use of LLMs as reviewers or meta-reviewers~\cite{du2024llms}, and proposed systems for summarizing or generating reviews~\cite{li2023summarizing, lin2024evaluating}. Together, these efforts highlight promising progress but remain fragmented, lacking a unified framework for comprehensive review quality assessment.
In parallel, several tools explicitly deploy LLMs to support review quality at scale. For example, the ICLR 2025 Review Feedback Agent provides structured, optional feedback on clarity, specificity, and professionalism to thousands of reviewers in a randomized study~\cite{thakkar2025llmfeedbackenhancereview}. Other systems explore automated peer-review generation and iterative review loops for academic writing~\cite{taechoyotin2025remorautomatedpeerreview}. We view \textsc{PeeriScope} as part of this emerging ecosystem rather than a stand-alone or definitive solution. \textsc{PeeriScope} offers an additional, complementary tool focused on post-hoc, multidimensional assessment of review helpfulness that can plug into existing reviewer training, monitoring, and decision-support workflows.
\textsc{PeeriScope} integrates structured linguistic metrics, LLM-based scoring, and supervised modeling to capture diverse aspects of review helpfulness. 
% Trained on expert-annotated reviews, it provides interpretable diagnostics and quantitative assessments through an accessible web interface and API. By combining interpretability with the power of foundation models, \textsc{PeeriScope} advances the development of trustworthy, automated review evaluation.

\begin{figure*}
    \centering
    \vspace{-1em}
    \includegraphics[width=1.77\columnwidth]{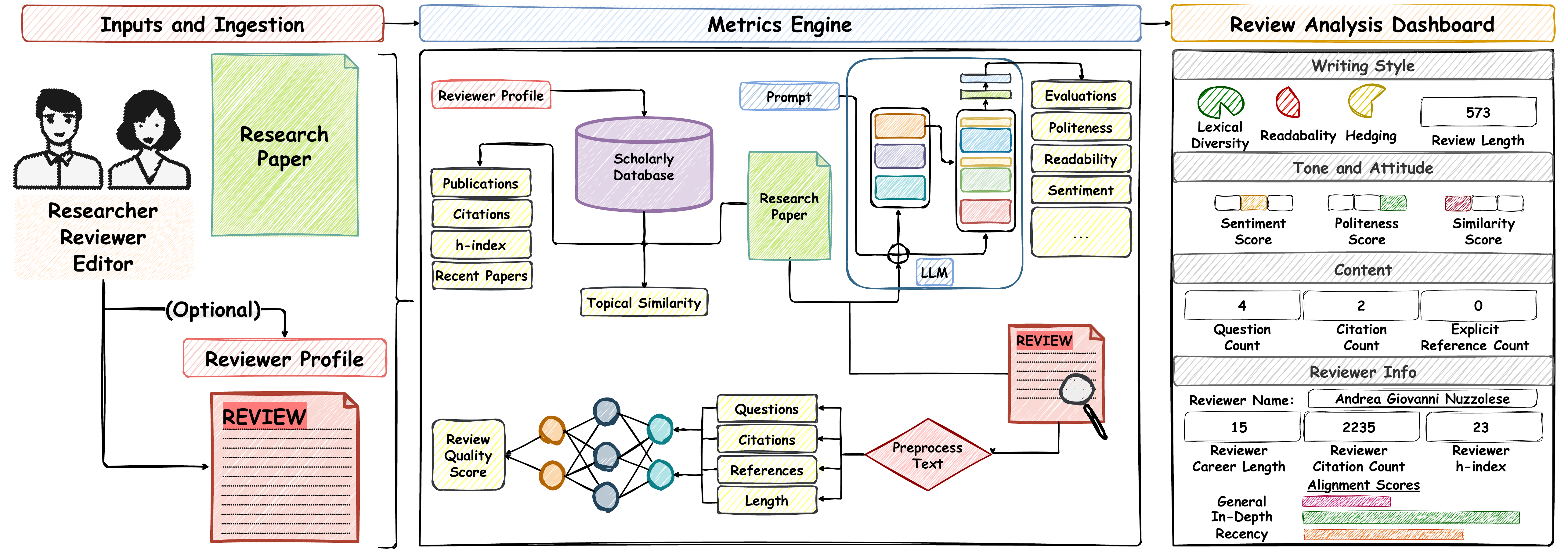}
        \vspace{-1em}
    \caption{Overview workflow of \textsc{PeeriScope}.}
        \vspace{-1em}
    \label{fig:overview}
\end{figure*}

\section{\textsc{PeeriScope} Overview}
\label{sec:overview}
\textbf{Design Requirements.}
Automated review quality assessment must satisfy both computational and workflow constraints. First, it must be \textit{scalable}. Conferences and journals handle thousands of submissions, requiring efficient, high-throughput evaluation. Second, it must be \textit{transparent} and \textit{interpretable} to editors, reviewers, and authors who rely on these signals for quality control and decisions. Finally, it must be \textit{compatible} with existing ecosystems and integrate smoothly with current infrastructures. Figure~\ref{fig:overview} summarizes the architecture of \textsc{PeeriScope} under these considerations.

\textbf{Inputs and Ingestion.} \label{sec:inputs-ingestion}
\textsc{PeeriScope} supports two input modes: In individual review mode, users provide the paper title, abstract, and full review text via a web form, intended for targeted editorial checks or reviewer self-evaluation. An optional OpenAlex identifier lets the system retrieve the reviewer’s publication profile and derive topical-expertise and citation features. 
%In OpenReview mode, users supply the URL of a public submission, and the system uses the OpenReview API to fetch manuscript metadata and all associated reviews. Both modes map inputs to a unified internal schema (Figure~\ref{fig:overview}) with structured fields for paper metadata, review texts, and optional reviewer profiles.

\textbf{Metrics Engine.}
\textsc{PeeriScope} evaluates peer review quality using three complementary groups of metrics: (1) structured metrics derived from the review text and optionally reviewer-profile metrics obtained from scholarly metadata, (2) rubric-guided LLM-based scores for abstract qualities such as constructiveness, and (3) a supervised overall quality estimator that integrates these signals into a single score. Further details on these metric categories are provided in Section~\ref{sec:interpretablemetrics}.
%The output of metric engine should approximate human judgments on review quality from different perspectives while preserving interpretability and computational efficiency.

\textbf{Review Analysis Dashboard.}
% \textsc{PeeriScope} exposes all signals from the metrics engine through an interactive dashboard with an overall quality bar, metric cards, and drill-down tabs that group outputs into interpretable categories, favoring transparency over a single opaque score and move beyond a single recommendation label (e.g., {weak accept}) toward a richer, multidimensional view of review quality. For settings where a web interface is not desirable, we additionally provide a programmatic API that returns structured JSON outputs for single reviews or batches, enabling automated quality auditing at scale.

\textsc{PeeriScope} exposes metrics through an interactive dashboard that emphasizes transparency over a single opaque score, offering a multidimensional view of review quality beyond simple recommendation labels. For non-interactive settings, it also provides a programmatic API that returns structured JSON outputs for single reviews or batches, enabling automated quality auditing at scale.

\vspace{-0.5em}
\section{\textsc{PeeriScope} Metrics}
\label{sec:metrics}

\textsc{PeeriScope} evaluates peer review quality using three complementary sources of evidence, introduced in the following subsections.

\vspace{-1em}
\subsection{Structured Metrics}
\label{sec:interpretablemetrics}

Our work is inspired by a strong line of prior works on assessment of scientific peer reviews and review helpfulness \cite{liu2023labels, ebrahimi2025rottenreviews, ebrahimi2025exharmony, arabzadeh2025building}, which has typically defined metrics around three classes: (i) writing style and readability ~\cite{xiong2011automatically,quaderi2024identification}, (ii) tone and reviewer attitude (e.g., sentiment, politeness)~\cite{bharti2024politepeer,sahakyan2025disparitiespeerreviewtone}, and (iii) the substantive content and coverage of the critique (e.g., section coverage, informativeness)~\cite{ramachandran2017automated,ghosal2022peer}.

\textbf{Writing Style} includes metrics related to reviewer effort and communicative clarity. \textit{Review length} is reported as a proxy for thoroughness. In addition, we report \textit{hedging} , which  marks uncertainty and plays an important role in balancing authority with caution. 
\textit{Hedging} is measured using a cue-based neural detector and captures the epistemic stance of the reviewer. 
\textit{Lexical diversity}, measured as type-token ratio, reflects linguistic variation and effort, and \textit{readability}, captured using the Flesch Reading Ease score, reflects how easily the review can be understood.

\textbf{Tone and Attitude} are modeled through \textit{politeness}, \textit{sentiment}, and \textit{similarity} between the review and the paper. \textit{Politeness} has been linked to perceptions of fairness and author receptivity, while \textit{sentiment} polarity offers a coarse but informative indicator of evaluative direction. The general \textit{similarity} of the review to the manuscript reflects the reviewer’s overall domain proximity. 

\textbf{Review Content} is also being assessed using metrics such as \textit{mentions of manuscript structure}; e.g., references to figures or specific sections, which suggest close reading and submission-specific critique. Additionally, we consider \textit{mentions of citations}, which provide support for reviewers’ claims and help ground them. Engagement is further captured through the \textit{presence of questions}. Constructive reviews often contain forward-looking or clarifying questions that invite reflection or revision. We identify these using a fine-tuned classifier trained on interrogative forms that indicate substantive reviewer intent. 

Textual features alone cannot capture the credibility or relevance of the reviewer. We therefore incorporate \textbf{reviewer-based metrics} using metadata from OpenAlex\footnote{\url{https://openalex.org/}}. We measure \textit{topical alignment} between the submission and the reviewer’s publication history using SPECTER~\cite{cohan2020specter}.
Reviewer standing is further characterized through \textit{citation count} and \textit{scholarly tenure}, a proxy for influence within a field. Together, these features provide complementary views of seniority, continuity, and visibility in the research community.

\subsection{LLM-based Metrics}
\label{sec:llmmetrics}

% Structured metrics provide interpretable signals of review quality from observable textual and contextual attributes. However, many important properties of peer reviews—such as fairness, constructiveness, factuality, and overall utility—are abstract and hard to reduce to surface-level proxies. To capture these dimensions, \textsc{PeeriScope} incorporates a second class of LLM-based evaluative signals. Prior work has shown that, when carefully prompted, LLMs can produce judgments that approximate expert annotations~\cite{gu2024survey}.

% We adopt Qwen-3-8B, a multilingual instruction-tuned LLM deployed locally, as our primary LLM-based evaluator to ensure data privacy and fast inference, though \textsc{PeeriScope} can readily swap in any comparable model. For each review, the LLM receives the full review text plus the title and abstract of the associated paper, and is prompted to score the review along several qualitative dimensions. Each dimension is paired with a rubric that specifies the scale and anchors scores in concrete criteria, adapted from editorial guidelines and prior review-quality annotation schemes~\cite{zahorodnii2025paper}. The full rubric set is given in Table~\ref{tab:metrics_kendall}. These scores are not meant to replace human judgment, but to provide a complementary layer that captures latent qualities beyond traditional features.

While structured metrics provide interpretable signals from observable textual and contextual attributes, many important aspects of review quality—such as fairness, constructiveness, factuality, and overall utility—are abstract and difficult to capture with surface-level proxies. To address these dimensions, \textsc{PeeriScope} incorporates LLM-based evaluative signals, building on prior work showing that carefully prompted LLMs can approximate expert judgments~\cite{gu2024survey}. We use Qwen-3-8B, a locally deployed instruction-tuned LLM, as the primary evaluator to ensure data privacy and fast inference, though the system remains model-agnostic. For each review, the model receives the full review text along with the title and abstract of the associated paper, and scores the review across multiple qualitative dimensions. Each dimension is paired with a rubric that defines the scoring scale and anchors judgments in concrete criteria, adapted from editorial guidelines and prior review-quality annotation schemes~\cite{zahorodnii2025paper}. The full rubric set is given in Table~\ref{tab:metrics_kendall}.

\begin{table*}[t]
\centering
\vspace{-1em}
\caption{Review quality dimensions and Kendall’s $\tau$ correlation between human and LLM judgments across three models.}
\vspace{-1em}
\label{tab:metrics_kendall}
\scalebox{0.8}{
\begin{tabular}{p{3.2cm} p{11cm} ccc}
\toprule
\textbf{Aspect} & \textbf{Description} & \textbf{GPT-4o} & \textbf{Phi-4} & \textbf{Qwen-3} \\
\midrule
Overall Quality         & Holistic evaluation of the review's usefulness and professionalism. & 0.359 & 0.241 & 0.252 \\
Comprehensiveness       & Covering all key aspects of the paper. & 0.476 & 0.374 & 0.338 \\
Actionability           & Helpfulness of the review in suggesting clear next steps. & 0.411 & 0.279 & 0.314 \\
Sentiment Polarity      & Overall sentiment conveyed by the reviewer. & 0.407 & 0.397 & 0.428 \\
Constructiveness        & Whether the review suggests improvements rather than only criticism. & 0.343 & 0.259 & 0.211 \\
Use of Technical Terms  & Using domain-specific vocabulary. & 0.327 & 0.254 & 0.176 \\
Objectivity             & Presence of unbiased, evidence-based commentary. & 0.298 & 0.215 & 0.186 \\
Alignment               & Relevance of the review to the scope of the paper. & 0.295 & 0.204 & 0.105 \\
Vagueness               & Degree of ambiguity or lack of specificity in the review. & 0.189 & 0.175 & 0.078 \\
Fairness                & Perceived impartiality and balance in judgments. & 0.163 & 0.186 & 0.139 \\
Politeness              & Tone and manner of the review language. & 0.128 & 0.053 & 0.106 \\
Clarity and Readability & Ease of understanding the review, including grammar and structure. & 0.124 & 0.038 & 0.117 \\
Factuality              & Accuracy of the statements made in the review. & 0.115 & 0.006 & 0.089 \\
\bottomrule
\end{tabular}}
\label{table:metrics}
\vspace{-1em}
\end{table*}

\subsection{Overall Quality Estimator}
\label{sec:learnt}

While structured and model-based metrics provide useful signals about specific dimensions of peer review quality, they do not by themselves yield a unified assessment that reflects how expert evaluators synthesize these attributes into an overall judgment. \textsc{PeeriScope} addresses this with a supervised scoring component trained on expert-annotated data to map heterogeneous features to a continuous quality score approximating human judgments.

We use two classes of models to estimate overall quality. The first class consists of lightweight regressors trained on the full set of structured and LLM-derived features. These include a linear regression model, a random forest, and a two-layer MLP. 
% Their simplicity offers three advantages: (i) They require no pretrained language model, which reduces computational overhead. (ii) They preserve feature-level interpretability, allowing users to trace model outputs back to specific input signals. (iii) They also support efficient inference, making them suitable for large-scale auditing tasks in conference or journal workflows.
The regressors' simplicity ensures low computational overhead, feature-level interpretability, and efficient inference suitable for large-scale auditing.
The second class of models incorporates a fine-tuned LLM that receives the title, abstract, and review text as input and is trained to regress directly to the human-annotated overall quality score. We use LLaMA3-8B with parameter-efficient fine-tuning based on low-rank adaptation and 8-bit weight quantization. We report agreement with human scores of both classes of models in \S\ref{sec:correlation}.

\section{Empirical Assessment of \textsc{PeeriScope}}
\label{sec:validation}

% We evaluate to what extend the quality evaluations produced by \textsc{PeeriScope} align with human expert judgments. We approach this by comparing the system's structured and LLM-based metrics, as well as its supervised quality estimators, against a set of reference annotations produced by trained human raters. 

% \noindent \textbf{Dataset.}
% We constructed a dataset of 753 peer reviews drawn from 200 papers published across OpenReview, F1000 Research, and the Semantic Web Journal. This data is available publicly on our HuggingFace for research purposes. Reviews were sampled to ensure coverage across venues and subject areas, with an emphasis on including both high- and low-quality examples. Each review was paired with the title and abstract of the corresponding paper and annotated independently by graduate students who were particularly trained for this task. Annotators scored each review along thirteen dimensions of quality, including comprehensiveness, fairness, clarity and more. An additional overall quality score was assigned using a continuous rubric designed to reflect editorial standards.

We assess how closely the quality evaluations produced by \textsc{PeeriScope} align with human expert judgments by comparing its structured, LLM-based, and supervised metrics against reference annotations from trained human raters.

\noindent \textbf{Dataset.}
We curated a dataset of 753 peer reviews from 200 papers across OpenReview, F1000 Research, and the Semantic Web Journal, released publicly for research use. Reviews were sampled to span venues, subject areas, and a range of quality levels. Each review was paired with the paper’s title and abstract and independently annotated by trained graduate students across thirteen quality dimensions. An additional overall quality score was assigned using a continuous rubric designed to reflect editorial standards.

\subsection{LLM-Human Alignment}
\label{sec:correlation}

To evaluate the degree to which \textsc{PeeriScope}'s rubric-aligned LLM outputs capture meaningful aspects of review quality, we compare model-generated scores with human annotations for each of the abstract dimensions included in the evaluation rubric. In each case, the LLM receives the review text along with the paper title and abstract and is prompted to assign quality scores on a five-point ordinal scale. We compute Kendall’s Tau correlation between the model’s ranked outputs and the corresponding human judgments.

Table~\ref{table:metrics} presents the correlation results for three models. GPT-4o serves as a high-capacity commercial baseline, while Phi-4 and Qwen-3 represent open-source alternatives. Among the evaluated models, GPT-4o achieves the highest alignment across most dimensions. However, absolute scores remain modest, with the highest observed correlation for overall quality reaching only 0.359. Several dimensions, such as factuality and clarity, show weak correlation across all models.
These findings are consistent with recent work in LLM-based evaluation \cite{li2025generationjudgmentopportunitieschallenges}, which has shown that zero-shot judgments, while often fluent and plausible, can fail to track ground-truth assessment in complex subjective tasks. In peer review, this limitation is pronounced, as criteria are context-dependent and LLMs capture broad patterns but often miss expert-level nuance.

\subsection{Comparison with Supervised Estimators}
We evaluate the overall quality estimators introduced in Section~\ref{sec:learnt} by measuring their alignment with human judgments. These include lightweight regressors trained on structured and LLM-based features, as well as a fine-tuned LLaMA3-8B model trained end-to-end on annotated review-paper pairs.

\begin{figure}
    \centering
    \vspace{-1em}
    \includegraphics[width=.8\columnwidth]{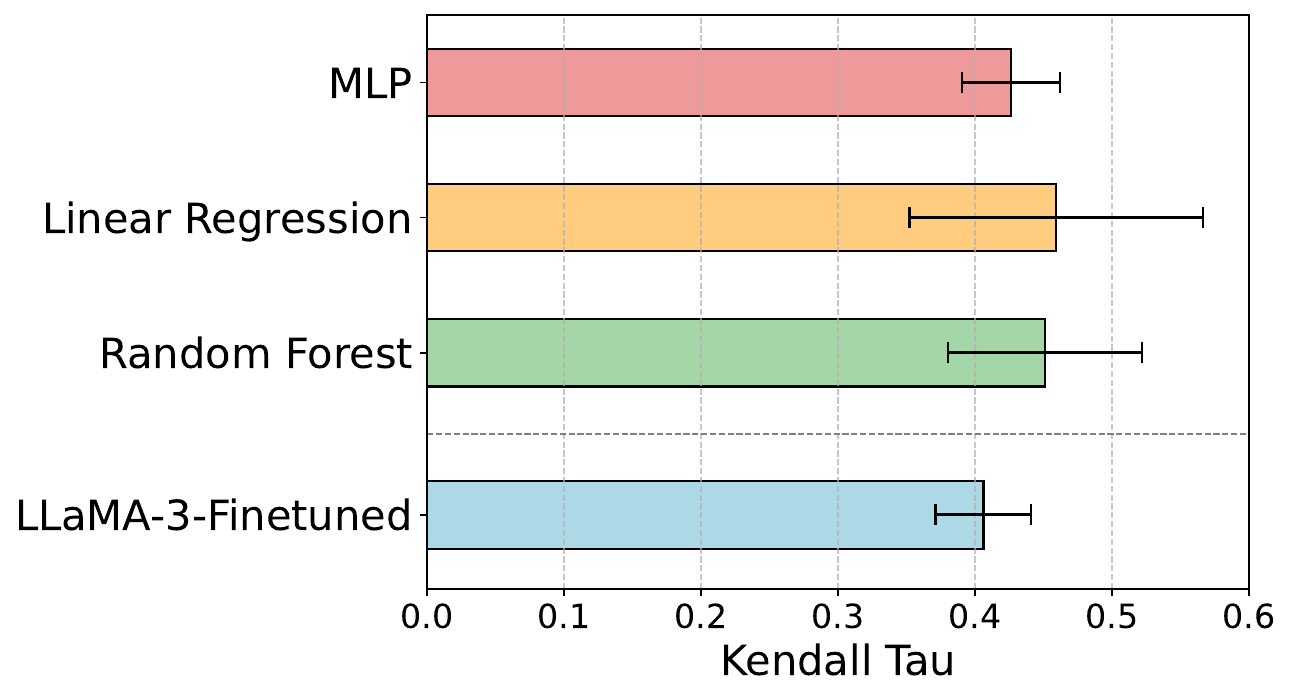}
        \vspace{-1em}
    \caption{Kendall’s $\tau$ correlation between human-evaluated and supervised overall quality estimators.}
        \vspace{-2em}
    \label{fig:kendal}
\end{figure}

Each model produces a scalar quality score for every review in the dataset. We compute Kendall’s Tau between the model predictions and human-assigned overall quality scores. Results on 10 fold cross validation are summarized in Figure~\ref{fig:kendal}, show that the structured-feature regressors outperform all zero-shot LLMs and also surpass the fine-tuned LLaMA model. Among these, a two-layer multilayer perceptron achieves the highest agreement, suggesting that relatively simple models can yield strong empirical performance when supported by carefully selected features.
These results offer several insights. First, structured features grounded in theory and linguistic analysis retain high predictive utility despite their simplicity. Second, fine-tuning large models on small-scale review datasets may not yield robust improvements, especially when evaluative reasoning depends on multiple latent dimensions not easily captured in training signals. Finally, the consistent outperformance of supervised predictors relative to zero-shot LLMs supports the use of hybrid systems like \textsc{PeeriScope}, which combine interpretable metrics, model-based evaluation, and supervised calibration.

\section{Implementation \& Deployment Details}
\label{app:implementation}

\textsc{PeeriScope} is built using a modular architecture that supports local and cloud deployment, with components implemented in FastAPI and React served by Docker containers. It exposes three REST endpoints for different review analysis tasks, integrates a locally hosted Qwen-3-8b via VLLM, and uses lightweight classifiers for complementary metrics. Reviewer metadata is sourced from OpenAlex and stored in MongoDB, with all computations performed in-memory to ensure privacy.
Incoming requests containing review or paper data are processed in-memory and discarded after metric computation. No content is stored or persisted beyond the duration of computation, thereby minimizing privacy exposure and satisfying lightweight compliance requirements.
The frontend client is implemented in React 18 with TypeScript and the Vite toolchain. It interacts with the backend through Axios, renders visual outputs using Recharts, and maintains session state in Redux Toolkit.
This architecture supports both standalone usage and embedded deployment in broader editorial systems.

\section{Concluding Remarks}
\label{sec:conclusion}
% This demonstration introduces \textsc{PeeriScope}, a system for evaluating peer-review quality using structured metrics, LLM-based assessments, and supervised prediction. Grounded in discourse theory and validated against expert annotations, \textsc{PeeriScope} supports reviewer self-assessment, editorial triage, and large-scale audit. It offers a practical tool for authors, reviewers, and organizers seeking greater transparency and accountability in scientific evaluation through interpretable signals and interactive exploration.
% The system is openly accessible via a web interface and API, with a modular design that supports extensions to new metrics and domains. We hope it contributes to shared infrastructure for improving peer review and advancing research on scholarly evaluation and feedback.

This demonstration introduces \textsc{PeeriScope}, a system for evaluating peer-review quality using structured metrics, LLM-based assessments, and supervised prediction. Validated against expert annotations, it supports reviewer self-assessment, editorial triage, and large-scale audit, providing interpretable signals and interactive exploration for greater transparency. The system is openly accessible via web interface and API, with a modular design for extending metrics and domains. We hope it contributes to infrastructure for improving peer review and advancing research on this topic.

\bibliographystyle{ACM-Reference-Format}
\bibliography{references}

\end{document}